\pdfoutput=1

\documentclass[11pt]{article}

\usepackage{emnlp2021}

\usepackage{times}
\usepackage{latexsym}
\usepackage{amsmath}
\usepackage{amssymb}
\usepackage{graphicx}
\usepackage{array}
\usepackage{booktabs, dcolumn, caption, bm}
\usepackage{multirow}
\usepackage{subcaption}
\usepackage{makecell}

\usepackage{float}
\usepackage{tabularx}
\newcolumntype{d}[1]{D{.}{.}{#1}} %
\newcommand{\bigcell}[2]{\begin{tabular}{@{}#1@{}}#2\end{tabular}}

\usepackage[T1]{fontenc}

\usepackage[utf8]{inputenc}

\usepackage{microtype}

\usepackage{enumitem}

\title{Towards Incremental Transformers: An Empirical Analysis of Transformer Models for Incremental NLU} 
\author{Patrick Kahardipraja \hspace{10mm} Brielen Madureira \hspace{10mm}  David Schlangen \\
  Computational Linguistics, Department of Linguistics \\ University of Potsdam, Germany \\
  \texttt{\{kahardipraja,madureiralasota,david.schlangen\}@uni-potsdam.de}}

\begin{document}
\maketitle
\begin{abstract} 
Incremental processing allows interactive systems to respond based on partial inputs, which is a desirable property \textit{e.g.} in dialogue agents. The currently popular Transformer architecture inherently processes sequences as a whole, abstracting away the notion of time. Recent work attempts to apply Transformers incrementally via \textit{restart-incrementality} by repeatedly feeding, to an unchanged model, increasingly longer input prefixes to produce partial outputs. However, this approach is computationally costly and does not scale efficiently for long sequences. In parallel, we witness efforts to make Transformers more efficient, \textit{e.g.}\ the Linear Transformer (LT) with a recurrence mechanism. In this work, we examine the feasibility of LT for incremental NLU in English. Our results show that the recurrent LT model has better incremental performance and faster inference speed compared to the standard Transformer and LT with restart-incrementality, at the cost of part of the non-incremental (full sequence) quality. We show that the performance drop can be mitigated by training the model to wait for right context before committing to an output and that training with input prefixes is beneficial for delivering correct partial outputs.
\end{abstract}

\section{Introduction}
One fundamental property of human language processing is \emph{incrementality} \citep{keller-2010-cognitively}. Humans process language on a word-by-word basis by maintaining a partial representation of the sentence meaning at a fast pace and with great accuracy \citep{marslen-wilson}. The garden path effect, for example, shows that language comprehension is approximated incrementally before committing to a careful syntactic analysis \citep[\textit{inter alia}]{FRAZIER1982178, ALTMANN1988191, TRUESWELL1994285}.

The notion of order along the time axis during computation is a key aspect of incremental processing and thus a desirable property both of cognitively plausible language encoders as well as in applications such as interactive systems \citep{skantze-schlangen-2009-incremental}. RNNs, for example, are inherently able to process words sequentially while updating a recurrent state representation. However, the Transformer architecture \citep{transformers}, which has brought significant improvements on several NLP tasks, processes the input sequence as a whole, thus prioritising parallelisation to the detriment of the notion of linear order.  

One way to employ non-incremental models in incremental settings is resorting to an incremental interface, like in \citet{, beuck-etal-2011-decision}, where a complete recomputation of the available partial input happens at each time step to deliver partial output. \citet{madureira-schlangen-2020-incremental} examined the output stability of non-incremental encoders in this \textit{restart-incremental} fashion. While qualitatively feasible, this procedure is computationally costly, especially for long sequences, since it requires as many forward passes as the number of input tokens.

In parallel, there is ongoing research on ways to make Transformers more efficient, \textit{e.g.} the Linear Transformer (LT) introduced by \citet{pmlr-v119-katharopoulos20a}. Besides being more efficient, LTs can be employed with a recurrence mechanism based on causal masking that turns them into models similar to RNNs. In this work, we examine the suitability of using LTs in incremental processing for sequence tagging and classification in English. We also inspect the use of the delay strategy \citep{baumannetal2011, oda-etal-2015-syntax, ma-etal-2019-stacl} to examine the effect of \textit{right context} availability on the model's incremental performance. Our hypothesis is that recurrence will allow LTs to be better in incremental processing as it captures sequence order. As LTs use an approximation of softmax attention, we also expect a performance drop compared to the standard Transformer while being faster in the incremental setting due to its linear time attention.

\section{Related Work}
In recent years, neural approaches, including Transformer-based architectures \citep{transformers}, have become more popular for incremental processing. Given that Transformer models are not inherently incremental, employing them for incremental processing demands adaptation. 

In simultaneous translation, for instance, \citet{ma-etal-2019-stacl} proposed to use an incremental encoder by limiting each source word to attend to its predecessors and recompute the representation for previous source words when there is new input. \citet{zhang2020futureguided} introduced an average embedding layer to avoid recalculation when using an incremental encoder, while exploiting \textit{right context} through knowledge distillation. An investigation of the use of non-incremental encoders for incremental NLU in interactive systems was conducted by \citet{madureira-schlangen-2020-incremental}. The authors employed BERT \citep{devlin-etal-2019-bert} for sequence tagging and classification using restart-incrementality, a procedure with high computational cost.  

The computational cost of a restart-incremental Transformer can be reduced with more efficient models or even avoided if an inherently incremental Transformer architecture existed. Recent works have proposed modifications that could help achieve that. For instance, by approximating the softmax attention with a recurrent state \citep{pmlr-v119-katharopoulos20a, choromanski2021rethinking, peng2021random}. The Linear Transformer model \citep[LT henceforth]{pmlr-v119-katharopoulos20a} can be viewed as an RNN when the attention is causal (see also, very recently, \citealp{kasai2021finetuning}).

\vspace*{-.2cm}
\section{Methods}
\vspace*{-.2cm}
\subsection{Overview of the Linear Transformer}
In LTs, the similarity score between a query and a key for the $i$-th position is computed using a kernel function. The causal attention can be written as:

\vspace*{-.3cm}
\begin{align}
    \text{Att}_{i}(Q, K, V) &= \frac{\phi (Q_{i})^{T}S_{i}}{\phi (Q_{i})^{T}Z_{i}} \\ 
    S_{i} = \sum_{j=1}^{i}\phi (K_{j})V_{j}^{T}&; Z_{i}= \sum_{j=1}^{i}\phi (K_{j})
\end{align}

with feature map $\phi (x) = \text{elu}(x) + 1$ where elu($\cdot$) denotes the exponential linear unit \citep{DBLP:journals/corr/ClevertUH15}. Hence, $S_{i}$ and $Z_{i}$ can be viewed as a recurrent state:

\vspace*{-.6cm}
\begin{align}
    S_{i} &= S_{i-1} + \phi(K_{i})V_{i}^{T} \\
    Z_{i} &= Z_{i-1} + \phi(K_{i})
\end{align}

with $S_{0} = Z_{0} = 0$. As an RNN, the run-time complexity is linear with respect to the sequence length and constant for each added token, which promises faster inference compared to the restart-incremental approach.

\subsection{Models}
We examine the behaviour of Transformer models used as incremental processors on token level, in five configurations (Table \ref{table:models}):

\begin{enumerate}[label=\textbf{\arabic*}.]
    \vspace{-0.2cm} \item \textbf{Baseline}: the standard Transformer encoder incrementalised via restart-incrementality, trained with access to full sequences.
    
    \vspace{-0.2cm} \item \textbf{LT}: the LT encoder incrementalised via restart-incrementality, trained with access to full sequences.
    
    \vspace{-0.2cm} \item \textbf{LT+R}: the LT encoder trained as in (2) but during test time we use its recurrent state vector to predict the label at each time step, as in an RNN.

    \vspace{-0.2cm} \item \textbf{LT+R+CM}: the LT encoder trained with causal masking to ensure each token representation can only attend to previous tokens. During inference, we convert the model to an RNN as in (3). Training with input prefixes aims at encouraging the learning of intermediate structures \citep{kohn-menzel-2014-incremental} and the anticipation of future output \citep{ma-etal-2019-stacl}.
    
    \vspace{-0.2cm} \item \textbf{LT+R+CM+D}: similar to (4), but, during training, the output for the input token $x_{t}$ is obtained at time $t+d$, where $d \in \{1, 2\}$ is the delay, following the approach in \citet{pmlr-v119-turek20a}. There is evidence that additional \textit{right context} improve the models' incremental performance
\citep{baumannetal2011, ma-etal-2019-stacl, madureira-schlangen-2020-incremental}, which results in a  trade-off between providing timely output or waiting for more context to deliver more stable output.
        
\end{enumerate}

We also delay the output by 1 and 2 time steps for the baseline and LT following \citet{madureira-schlangen-2020-incremental}, to provide a fair comparison on incremental metrics. Note that outputs from both (1) and (2) are non-monotonic, as labels can be reassigned when a new input token is observed. The other models deliver monotonic output for sequence tagging as RNNs. A slight modification is needed for sequence classification as each sequence is mapped to a single label. We average the hidden representation at the last layer and project it linearly, followed by a softmax to obtain the sequence label $\hat{y}_{t}$ based on the consumed input until time $t$. For LT+ models, we use incremental averaging to avoid recomputation. By doing this, sequence classification is performed similarly for all models.

\begin{table}[h] \small
\captionsetup{singlelinecheck = false, justification=justified}
\setlength\tabcolsep{2.25pt}
\begin{tabular*}{\columnwidth}{l c c c c}
\toprule
Models & \bigcell{c}{Restart \\ Incremental} & Recurrence & \bigcell{c}{Causal \\ Masking} & \multicolumn{1}{c}{Delay} \\ 
\midrule
Baseline & \checkmark & - & -  & \multicolumn{1}{c}{-*}  \\
LT & \checkmark & - & -  & -*  \\
LT+R & - & \checkmark & -  &  -\hphantom{*}  \\
LT+R+CM & - & \checkmark & \checkmark  & -\hphantom{*}  \\
LT+R+CM+D & - & \checkmark & \checkmark  & \checkmark\hphantom{*}  \\
\bottomrule
\end{tabular*}
\caption{Overview of the Transformer models. * means we perform further comparisons with a delayed variant.}
\label{table:models}
\end{table}

\vspace*{-.3cm}
\section{Experimental Setup}
\vspace*{-.1cm}

\subsection{Datasets}
We evaluate our models on 9 datasets in English, which were also used in \citet{madureira-schlangen-2020-incremental}. The tasks consist of sequence tagging: slot filling (ATIS, \citet{hemphill-etal-1990-atis,dahl-etal-1994-expanding} and SNIPS, \citet{coucke2018snips}), chunking (CoNLL-2000, \citet{tjong-kim-sang-buchholz-2000-introduction}), NER and PoS tagging (OntoNotes 5.0, WSJ section, \citet{ontonotes}); and sequence classification: intent detection (ATIS and SNIPS) and sentiment classification (positive/negative, \citet{10.1145/2783258.2783380} and pros/cons, \citet{ganapathibhotla-liu-2008-mining}). More details are available in the Appendix.

\vspace*{-.1cm}
\subsection{Evaluation}
\vspace*{-.1cm}
The overall performance of the models is measured with accuracy and F1 Score, according to the task. For the incremental evaluation, we report the diachronic metrics proposed by \citet{baumannetal2011} and adapted in \citet{madureira-schlangen-2020-incremental}: \emph{edit overhead} (EO, the proportion of unnecessary edits over all edits), \emph{correction time score} (CT, the average proportion of time steps necessary to reach a final decision), and \emph{relative correctness} (RC, the proportion of time steps in which the output is a correct prefix of the final, non-incremental output). 

To focus on the incremental quality of the models and allow a clear separation between incremental and non-incremental evaluation, we follow the approach by \citet{baumannetal2011} and \citet{madureira-schlangen-2020-incremental}, evaluating incremental outputs with respect to the final output produced by the models. While the final output may differ from the gold standard, it serves as the target for the incremental output, as the non-incremental performance is an upper bound for incremental processing \citep{baumannetal2011}.

\vspace*{-.1cm}
\subsection{Implementation}
\vspace*{-.1cm}
We re-implement the Transformer and use the original implementation of the LT.\footnote{\url{https://linear-transformers.com}} All models are trained to minimise cross-entropy with the AdamW optimiser \citep{loshchilov2018decoupled}. We use 300-D GloVe embeddings \citep{pennington-etal-2014-glove} which are passed through a linear projection layer with size $d_{\text{model}}$. All experiments were performed on a GPU GeForce GTX 1080 Ti. Details on the implementation, hyperparameters and reproducibility are available in the Appendix. Our implementation is publicly available.\footnote{\url{https://github.com/pkhdipraja/towards-incremental-transformers}}
\vspace*{-.1cm}

\section{Results and Discussion}
\vspace*{-.3cm}

\begin{table}[ht] 
{\footnotesize
\captionsetup{singlelinecheck = false, justification=justified}
\setlength\tabcolsep{3.5pt}
\label{turns}
\hspace*{-.12in}
\begin{tabular}{l @{} c @{\;} c c c @{\;} c @{\;} c}
\toprule
Tasks & Baseline & LT & LT+R & \bigcell{c}{LT+R \\ +CM} & \bigcell{c}{LT+R+ \\ CM+D1 } & \bigcell{c}{LT+R+ \\ CM+D2}\\
\midrule
ATIS-Slot     & \textbf{94.51} & 93.67 & 86.84 & 93.78 & 94.38 & 93.54 \\
SNIPS-Slot    & \textbf{90.13} & 87.98 & 63.16 & 81.88 & 85.72 & 86.91   \\
Chunk    & \textbf{91.27} & 88.42 & 67.54 & 86.63 & 89.42 & 89.33   \\
NER    & \textbf{89.55} & 86.13 & 52.04 & 69.09 & 81.39 & 85.55   \\
\midrule
PoS Tagging  & \textbf{96.88} & 96.49 & 89.30 & 95.11 & 96.49 & 96.55    \\
ATIS-Intent  & \textbf{97.20} & 97.09 & 95.63 & 95.74 & 96.53 & 96.53   \\
SNIPS-Intent & \textbf{97.14} & \textbf{97.14} & 83.71 & 96.43 & \textbf{97.14} & 96.86   \\
Pos/Neg & \textbf{86.00} & 85.17 & 68.00 & 80.33 & 81.16 & 82.67  \\
Pros/Cons   & 94.42 & 94.21 & 90.97 & 94.37 & \textbf{94.56} & 94.46   \\

\bottomrule
\end{tabular}
\vspace*{-.2cm}
\caption{Non-incremental performance of our models on test sets (first group, F1, second group, accuracy). Here, the baseline performs generally better than the LT variants.}
\label{table:nonincremental}
}
\vspace*{-.3cm}
\end{table}

Ultimately, the quality of the output when all input has been seen matters; hence, we first look at the \textit{non-incremental} or full-sequence performance, in Table \ref{table:nonincremental}. We can see that LTs do not outperform the baseline here,\footnote{Notice that the baseline differs from \citet{madureira-schlangen-2020-incremental} who used a pretrained BERT model.} although they have the advantage of being faster (Table \ref{table:benchmark}). We see two possible reasons for this: First, as the LT variants strictly go left-to-right through the sequence, they have less information for each token to make their decision.
This can be alleviated by allowing LT+R+CM to wait for 1 or 2 tokens before producing partial output, and indeed we see an overall performance increase of 0.1\% - 16.5\% for the +D variants. Second, we suspect that the chosen feature map in LTs to approximate the softmax attention is sub-optimal, and a further gating mechanism could yield a better performance \citep{peng2021random}.

Comparing LT+R+CM against LT+R, we observe that training on prefixes yields a better result during test time, as LT+R+CM may learn anticipation as a by-product, in line with the work of \citet{ma-etal-2019-stacl}. LT+R+CM+D performs competitively with LT, outperforming the latter in 4 out of 9 datasets. This is likely due to the better capability of the delayed network in modelling both non-linear and acausal functions that appear in some of the tasks \citep{pmlr-v119-turek20a}.

\begin{figure}[h!]
\centering
\begin{subfigure}[b]{0.5\textwidth}
   \centering
   \includegraphics[width=0.9\linewidth]{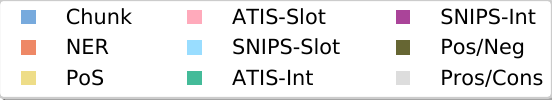}
\end{subfigure}
\begin{subfigure}[b]{0.5\textwidth}
   \centering
   \includegraphics[width=0.9\linewidth]{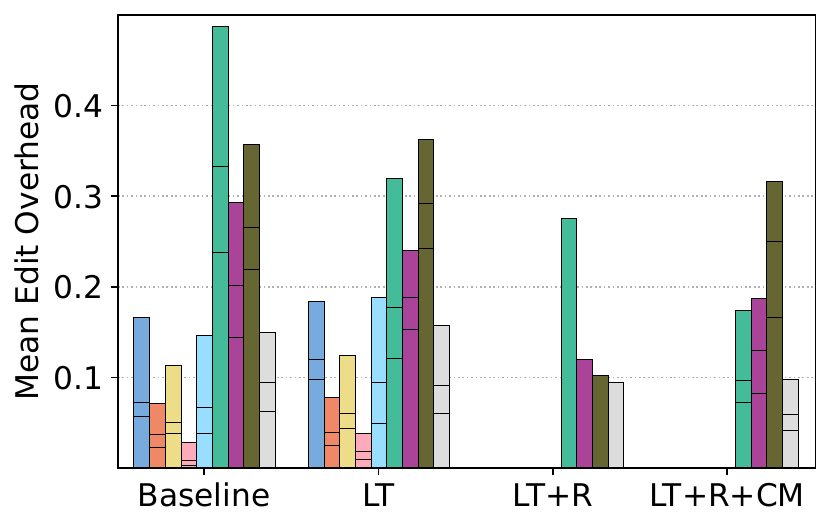}
   \vspace*{-.3cm}
   \vspace*{.3cm}
   \label{fig:meanEO} 
\end{subfigure}
\begin{subfigure}[b]{0.5\textwidth}
   \centering
   \includegraphics[width=0.9\linewidth]{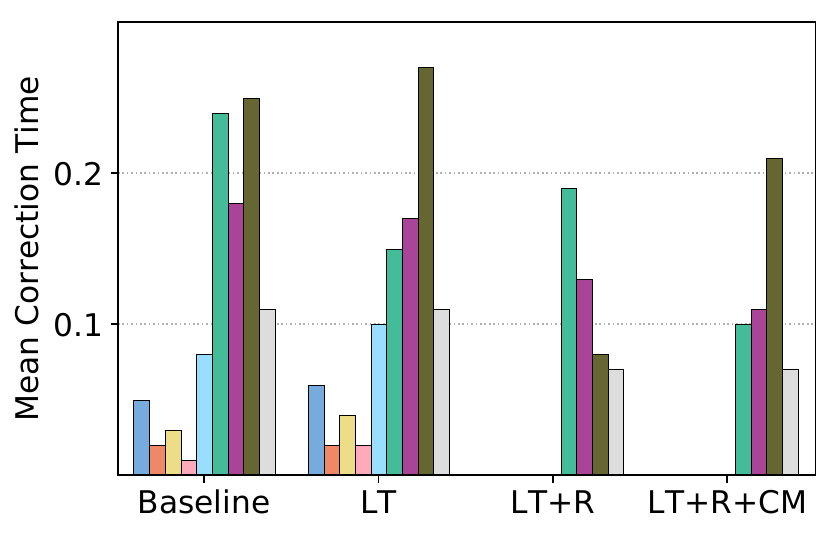}
    \vspace*{-.3cm}
    \vspace*{.3cm}
\label{fig:meanCT} 
\end{subfigure}
\begin{subfigure}[b]{0.5\textwidth}
   \centering
   \includegraphics[width=0.9\linewidth]{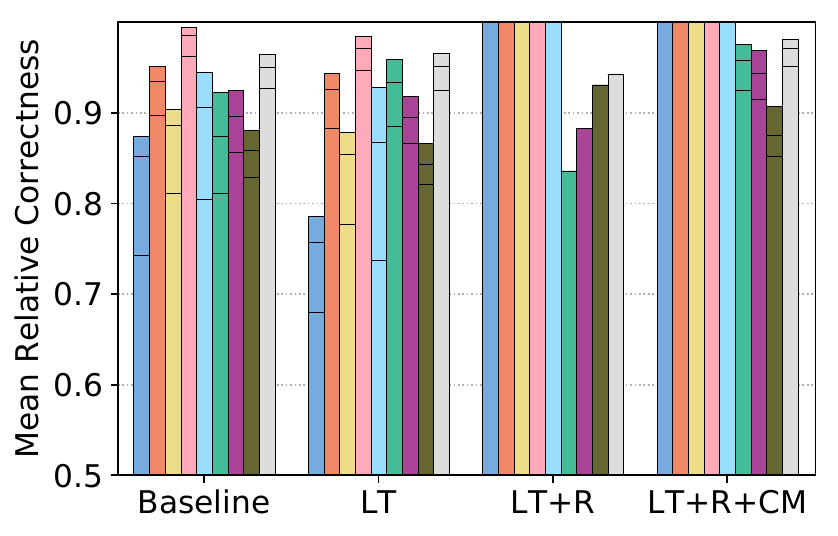}
\vspace*{-.3cm}
   \label{fig:meanRC} 
\end{subfigure}
  \caption{Incremental evaluation on the test sets. EO, CT and RC $\in$ [0, 1], y-axes are clipped to improve readability. Lower is better for EO and CT, higher for RC. For EO, the lines on the bars refer to original, delay=1 and delay=2, from top to bottom, and vice versa for RC, showing that delay improves the results. LT+R+CM performs better compared to the baseline and LT.}
  \label{fig:incrementalmetrics} 
\end{figure}

Figure \ref{fig:incrementalmetrics} depicts the incremental metrics of all models. The EO and CT score for sequence tagging is low for all models, which indicates that the models are capable, in general, of producing stable and accurate partial outputs. Notice that the LT+ models are not able to revise the output in sequence tagging. For sequence classification, the models have higher EO and CT score due to the fact that the label is a function of the whole sequence and the model might be unable to reach an early decision without enough right context. 

LT+R+CM performs better in incremental metrics compared to the baseline and LT in sequence classification. This is evidence that the notion of order is important for incremental processing, as the recurrent state in the LT allows partial representation updates along the time axis when processing partial input. Here, sequence classification is treated in a similar fashion for all models, by using the average of the hidden representation in the last layer. 

All the models have high RC score in general for both sequence tagging and classification. This means that most of the partial outputs are a correct prefix of the final (``non-incremental'') output and could fruitfully be used as input to subsequent processors in an incremental pipeline. For RC, LT+R+CM also outperforms both the baseline and LT in all tasks. A delay of 1 or 2 tokens before committing to an output also helps to improve the incremental performance across all models. In terms of incremental inference speed, we see that the recurrent mode is more than 10 times faster compared to using restart-incrementality (Table \ref{table:benchmark}).

\begin{figure}[h]
	\centering
	\includegraphics[width=\columnwidth]{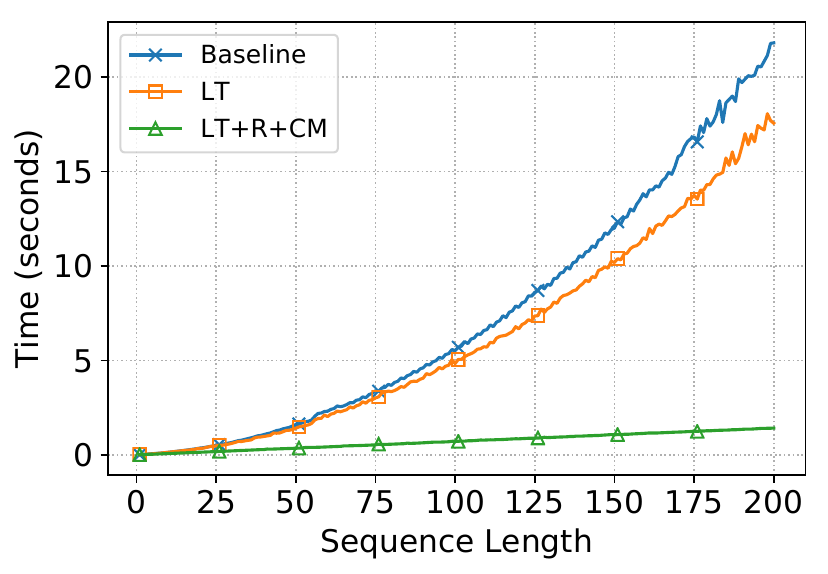}
	\caption{Incremental inference speed of models from Table \ref{table:benchmark} with increasing sequence length. LT+R+CM scales linearly with sequence length unlike the baseline and LT. Note that the incremental inference speed of LT+R+CM is similar to LT+R.}
	\label{fig:diff_sent_len}
\end{figure}

To understand the models' behaviour better, especially pertaining to their potential for real-time applications, we also examine their incremental inference speed for different sequence lengths as shown in  Figure \ref{fig:diff_sent_len}. As expected, LT+R+CM scales linearly and outperforms the baseline and LT considerably as the sequence becomes longer. The run-time performance of LT is slightly better than the baseline because of its linear time attention, however it is still slower compared to LT+R+CM as it is restart-incremental.

\begin{table}[h!] \small
\captionsetup{singlelinecheck = false, justification=justified}
\setlength\tabcolsep{0pt}
\begin{tabular*}{\columnwidth}{@{\extracolsep{\fill}} l c c c @{\hskip 0.25cm} r}
\toprule
 \multicolumn{1}{l}{Tasks} & \multicolumn{1}{c}{Baseline} & \multicolumn{1}{c}{LT} & \multicolumn{1}{c}{LT+R+CM} \\
\midrule
ATIS-Slot     & 0.983 & 1.025 & 13.780   \\
SNIPS-Slot    & 1.021 & 1.137 & 14.957  \\
Chunk   	  & 0.393 & 0.448 & \hphantom{1}6.023   \\
NER 		  & 0.382 & 0.436 & \hphantom{1}5.745    \\
PoS Tagging   & 0.383  & 0.435 & \hphantom{1}5.831   \\
ATIS-Intent   & 0.883  & 1.005 & 13.310   \\
SNIPS-Intent  & 0.995  & 1.129 & 14.907   \\
Pos/Neg 	  & 0.725  & 0.767 & \hphantom{1}9.962  \\
Pros/Cons 	  & 1.073  & 1.228 & 14.979   \\
\midrule
Average & 0.76 (1$\times$) & 0.85 (1.12$\times$) & \textbf{11.06} (\textbf{14.55}$\times$) \\
\bottomrule
\end{tabular*}
\caption{Comparison of incremental inference speed on test sets, measured in sequences/sec. All the models have similar size with 4 layers, feed-forward dimension of 2048 and self-attention dimension of 512.}
\label{table:benchmark}
\end{table}

\subsection{Ablations}
We examine the importance of word and positional embeddings on the baseline and LT+R+CM for non-incremental metrics (Table \ref{table:ablation}). We find that using pre-trained GloVe \citep{pennington-etal-2014-glove} embeddings is beneficial for the models' performance. On average, it contributes 2.74 accuracy and 5.16 F1 for the baseline, while improving LT+R+CM by 1.6 and 2.55 points for accuracy and F1. On the other hand, we observe that positional embeddings play a less significant role in LT+R+CM compared to the baseline. Without them the performance, on average, for LT+R+CM improves in accuracy by 0.15 and the F1 score degrades by 1.35. The baseline, however, experiences degradation in performance by 1.79 and 18.46 points for accuracy and F1, on average. The recurrence mechanism may be a reason for the effect of positional embeddings being less pronounced in LT+R+CM.

\vfill
\null
\begin{table}[h] 
{\footnotesize
\captionsetup{singlelinecheck = false, justification=justified}
\setlength\tabcolsep{4pt}
\begin{tabular*}{\columnwidth}{l l l r r r}
\toprule
& Tasks & Score & -- GloVe & -- Pos & \bigcell{r}{-- Glove \\ \& Pos}\\
\midrule
& \textbf{Baseline} \\
\cmidrule{2-2}
\multirow{4}{*}{\rotatebox[origin=c]{90}{\textbf{F1}}}
& \text{ATIS-Slot}     &  94.51 & -3.02 & -17.99 & -20.44 \\
& \text{SNIPS-Slot}    &  90.13 & -6.74 & -16.39 & -21.47 \\
& \text{Chunk}   	  &  91.27 & -5.28 & -18.19 & -19.36 \\
& \text{NER} 		  &   89.55 & -5.61 & -21.27 & -26.07 \\
\midrule
\multirow{4}{*}{\rotatebox[origin=c]{90}{\textbf{Accuracy}}}
& \text{PoS Tagging}   &  96.88 & -0.79 & -3.31 & -4.15 \\
& \text{ATIS-Intent}   &  97.20 & -2.35 & -2.69 & -3.81 \\
& \text{SNIPS-Intent}  &  97.14 & -0.57 & +0.72 & -0.43 \\
& \text{Pos/Neg} 	  &  86.00 & -8.83 & -3.83 & -12.00 \\
& \text{Pros/Cons} 	  &  94.42 & -1.18 & +0.15 & -1.64 \\
\midrule
& \textbf{LT+R+CM} \\
\cmidrule{2-2}
\multirow{4}{*}{\rotatebox[origin=c]{90}{\textbf{F1}}}
& \text{ATIS-Slot}     & 93.78 & -0.42 & -0.25 & -1.37 \\
& \text{SNIPS-Slot}    & 81.88  & -2.44 & -0.55 & -2.79 \\
& \text{Chunk}   	  & 86.63 & -3.82 & -3.35 & -7.30 \\
& \text{NER} 		  & 69.09 & -3.52 & -1.24 & -6.69  \\
\midrule
\multirow{4}{*}{\rotatebox[origin=c]{90}{\textbf{Accuracy}}}
& \text{PoS Tagging}   & 95.11  & -0.74 & -0.79 & -1.58 \\
& \text{ATIS-Intent}   & 95.74  & -0.89 & -0.11 & -0.67 \\
& \text{SNIPS-Intent}  & 96.43  & +0.14 & 0.00 & +0.28 \\
& \text{Pos/Neg} 	  & 80.33  & -6.00 & +1.17 & -9.00\\
& \text{Pros/Cons} 	  & 94.37  & -0.53 & +0.46 & -0.83 \\
\bottomrule
\end{tabular*}
\caption{Ablation of GloVe and positional embeddings on the baseline and LT+R+CM for non-incremental metrics.}
\label{table:ablation}
}
\vspace*{-.4cm}
\end{table}

\section{Conclusion}
\vspace*{-.1cm}
We studied the use of Transformer encoders for incremental processing and concluded that it is possible to deploy them as incremental processors with certain trade-offs. With recurrent computation, the Linear Transformer (LT) has inferior non-incremental performance compared to the regular Transformer and the LT with restart-incrementality. However, it has the great advantage of being much more efficient for incremental processing, since recomputation at each time step is avoided. The output of the recurrent LT is generally more stable for sequence classification and monotonic for tagging. Its non-incremental performance drop can be mitigated by introducing delay, which also improves the incremental metrics. It is also beneficial to train such model with input prefixes, allowing it to learn more robust predictions.

\section*{Acknowledgements}
\vspace*{-.1cm}
We thank the anonymous reviewers for their critical reading of our manuscript and their insightful comments and suggestions. This work is partially funded by the Deutsche Forschungsgemeinschaft (DFG, German Research Foundation) -- Project ID 423217434 (Schlangen).

\section*{Corrigendum}
Due to a bug in the implementation of delayed EO, our results on incremental metrics were slightly less accurate in the previous version. We have updated the results in Figure \ref{fig:incrementalmetrics}, Table \ref{table:seqtagging} and Table \ref{table:seqclassification} to reflect the changes. The text in the main paper stays the same, since no conclusions changed.

\vspace*{-.3cm}
\bibliography{anthology,custom}
\bibliographystyle{acl_natbib}

\newpage
\appendix

\section{Reproducibility}
\label{sec:reproducibility}
We describe in more detail the hyperparameters and implementation of our models.

\section*{Data} 
We mostly follow \citet{madureira-schlangen-2020-incremental}. We use only the WSJ section of OntoNotes with splits following \citet{pradhan-etal-2013-towards}. For Pos/Neg and Pros/Cons datasets, we split them randomly with a proportion of 70\% train, 10\% validation, and 20\% test set due to the unavailability of an official splitting scheme. We removed sentences longer than 200 words as they were infeasible to compute. We use the preprocessed data and splits for SNIPS and ATIS made available by \citet{e-etal-2019-novel}.

\section*{Training details} 
Our models are trained for 50 epochs, using early stopping with patience of 10 and dropout of 0.1. For AdamW \citep{loshchilov2018decoupled}, we use $\beta_{1} = 0.9$ and $\beta_{2} = 0.98$. The learning rate is increased linearly for the first 5 epochs. After 30, 40, and 45 epochs, we decay the learning rate by 0.5. Xavier initialisation \citep{pmlr-v9-glorot10a} is applied to all parameters. The number of attention heads is set to 8, where the dimension of each head is self-attention dimension $d/8$.

We also apply label smoothing \citep{labelsmoothing} with $\epsilon = 0.1$ for sequence classification to make the model more robust for incremental processing. For OOV words, we randomly replace tokens by "UNK" token with $p= 0.02$ during training and use it for testing \citep{10.1007/978-3-319-24033-6_20}. We perform hyperparameter search using Comet's Bayesian search algorithm \footnote{\url{https://www.comet.ml/docs/python-sdk/introduction-optimizer/}}, maximising F1 score for sequence tagging and accuracy for sequence classification on the validation set. The hyperparameter search trials are limited to 20 for all of our experiments. The hyperparameters for LT were also used for LT+R. We use similar hyperparameters for LT+R+CM and LT+R+CM+D. We set the seed to 42119392 for all of our experiments.

\begin{table}[h] \small
\captionsetup{singlelinecheck = false, justification=justified}
\setlength\tabcolsep{10pt}
\begin{tabular*}{\columnwidth}{l l}
\toprule
 \multicolumn{1}{l}{Hyperparameters} & \\
\midrule
Layers     & 1, 2, 3, 4  \\
Gradient clipping    & \text{no clip}, 0.5, 1  \\
Learning rate &  $5e^{-5}, 7e^{-5}, 1e^{-4}$  \\
Batch size 		  & 32, 64, 128   \\
Feed-forward dimension   & 1024, 2048    \\
Self-attention dimension   & 256, 512    \\
\bottomrule
\end{tabular*}
\caption{Hyperparameter search space. We use the same search space for all of our models.}
\label{table:hyperparameter}
\end{table}

\begin{table}[h] \small
\captionsetup{singlelinecheck = false, justification=justified}
\setlength\tabcolsep{12pt}
\begin{tabular*}{\columnwidth}{l c}
\toprule
\multicolumn{1}{l}{Tasks} & \multicolumn{1}{c}{Average sequence length} \\
\midrule
ATIS-Slot &  10.26 \\
SNIPS-Slot & \hphantom{1}9.08 \\
Chunk & 23.55 \\
NER & 24.14 \\
PoS Tagging & 24.14 \\
ATIS-Intent & 10.26 \\
SNIPS-Intent & \hphantom{1}9.08 \\
Pos/Neg & 13.95 \\
Pros/Cons & \hphantom{1}8.99 \\
\bottomrule
\end{tabular*}
\caption{Average sequence length on test sets for each task.}
\label{table:avg_seq_len}
\vspace*{+15.5cm}
\end{table}

\begin{table*}[h] \footnotesize
\captionsetup{singlelinecheck = false, justification=justified}
\setlength\tabcolsep{3pt}
\begin{tabular*}{\textwidth}{@{\extracolsep{\fill}} l c c c c c c c}
\toprule
\multicolumn{1}{l}{Tasks} & \multicolumn{1}{l}{Layers} & \multicolumn{1}{l}{Gradient clip} & \multicolumn{1}{l}{Learning rate} & \multicolumn{1}{l}{Batch size} & \multicolumn{1}{l}{Feed-forward} & \multicolumn{1}{l}{Self-attention} \\
\midrule 
Baseline \\
\cmidrule{1-1}
ATIS-Slot    & 2 & \text{no clip} & $1e^{-4}$ & 32 & 1024 & 256 \\
SNIPS-Slot   & 3 & 0.5 & $1e^{-4}$ & 64 & 1024 & 512   \\
Chunk    & 4 & 1 & $7e^{-5}$ & 32 & 2048 & 512   \\
NER    & 4 & \text{no clip} & $7e^{-5}$ & 32 & 2048 & 512   \\
PoS Tagging   & 4 & 0.5 & $1e^{-4}$ & 32 & 1024 & 512    \\
ATIS-Intent   & 4 & \text{no clip} & $1e^{-4}$ & 32 & 1024 & 256   \\
SNIPS-Intent  & 3 & 0.5 & $1e^{-4}$ & 64 & 1024 & 512  \\
Pos/Neg  & 2 & 0.5 & $5e^{-5}$ & 64 & 2048 & 256  \\
Pros/Cons    & 3 & 1 & $7e^{-5}$ & 64 & 2048 & 512   \\

\midrule

LT \\
\cmidrule{1-1}
ATIS-Slot    & 3 & 1 & $7e^{-5}$ & 32 & 2048 & 512 \\
SNIPS-Slot    & 3 & \text{no clip} & $1e^{-4}$ & 32 & 1024 & 512   \\
Chunk    & 3 & 0.5 & $1e^{-4}$ & 64 & 1024 & 512   \\
NER    & 2 & 1 & $1e^{-4}$ & 32 & 1024 & 512   \\
PoS Tagging  & 2 & 0.5 & $7e^{-5}$ & 32 & 2048 & 512    \\
ATIS-Intent  & 2 & 1 & $7e^{-5}$ & 32 & 2048 & 512   \\
SNIPS-Intent  & 3 & 0.5 & $1e^{-4}$ & 32 & 1024 & 256  \\
Pos/Neg & 4 & 1 & $1e^{-4}$ & 64 & 2048 & 512  \\
Pros/Cons  & 3 & \text{no clip} & $7e^{-5}$ & 32 & 2048 & 256   \\

\midrule

LT+R+CM \\
\cmidrule{1-1}
ATIS-Slot     & 3 & \text{no clip} & $1e^{-4}$ & 64 & 2048 & 512 \\
SNIPS-Slot    & 3 & \text{no clip} & $7e^{-5}$ & 32 & 1024 & 512   \\
Chunk    & 3 & 1 & $7e^{-5}$ & 32 & 1024 & 512   \\
NER    & 4 & 1 & $7e^{-5}$ & 64 & 1024 & 512   \\
PoS Tagging  & 2 & \text{no clip} & $1e^{-4}$ & 64 & 1024 & 512    \\
ATIS-Intent  & 3 & 1 & $5e^{-5}$ & 32 & 2048 & 512   \\
SNIPS-Intent  & 3 & \text{no clip} & $1e^{-4}$ & 32 & 2048 & 512  \\
Pos/Neg  & 4 & 1 & $1e^{-4}$ & 32 & 2048 & 256  \\
Pros/Cons   & 2 & 0.5 & $5e^{-5}$ & 32 & 1024 & 512   \\

\bottomrule
\end{tabular*}
\caption{Hyperparameters used for our experiments. The best configuration for LT was also used for LT+R, while the best configuration for LT+R+CM was also used for LT+R+CM+D1 and LT+R+CM+D2.}
\end{table*}

\begin{table*}[h] \footnotesize
\captionsetup{singlelinecheck = false, justification=justified}
\setlength\tabcolsep{0pt}
\begin{tabular*}{\textwidth}{@{\extracolsep{\fill}} l c c c c c c c}
\toprule
\multicolumn{1}{l}{Tasks} & \multicolumn{1}{c}{Baseline} & \multicolumn{1}{c}{LT}
& \multicolumn{1}{c}{LT+R} & \bigcell{c}{LT+R \\ +CM} & \bigcell{c}{LT+R+ \\ CM+D1} & \bigcell{c}{LT+R+ \\ CM+D2}  \\
\midrule
ATIS-Slot     & \hphantom{1}\text{2.0M} & \hphantom{1}\text{9.9M} & \hphantom{1}\text{9.9M} & \hphantom{1}\text{9.9M} & \hphantom{1}\text{9.9M} & \hphantom{1}\text{9.9M}  \\
SNIPS-Slot    & \text{10.0M} & \text{10.0M} & \text{10.0M} & \text{10.0M} & \text{10.0M} & \text{10.0M}\\
Chunk   & \text{18.5M} & \text{12.2M} & \text{12.2M} & \text{12.2M} & \text{12.2M} & \text{12.2M}   \\
NER    & \text{24.4M} & \text{16.0M} & \text{16.0M} & \text{20.2M} & \text{20.2M} & \text{20.2M}   \\
PoS Tagging  & \text{20.2M} & \text{18.1M} & \text{18.1M}  & \text{16.0M} & \text{16.0M} & \text{16.0M}   \\
ATIS-Intent  & \hphantom{1}\text{3.5M} & \hphantom{1}\text{6.7M} & \hphantom{1}\text{6.7M} & \hphantom{1}\text{9.9M} & \hphantom{1}\text{9.9M} & \hphantom{1}\text{9.9M}   \\
SNIPS-Intent & \text{10.0M} & \hphantom{1}\text{6.0M} & \hphantom{1}\text{6.0M} & \text{13.1M} & \text{13.1M} & \text{13.1M}   \\
Pos/Neg & \hphantom{1}\text{4.3M} & \text{14.4M} & \text{14.4M} & \hphantom{1}\text{6.9M} & \hphantom{1}\text{6.9M} & \hphantom{1}\text{6.9M}  \\
Pros/Cons   & \text{14.1M} & \hphantom{1}\text{8.5M} & \hphantom{1}\text{8.5M} & \hphantom{1}\text{8.8M} & \hphantom{1}\text{8.8M} & \hphantom{1}\text{8.8M}   \\

\bottomrule
\end{tabular*}
\caption{Number of parameter for each model.}
\end{table*}

\begin{table*}[h] \footnotesize
\captionsetup{singlelinecheck = false, justification=justified}
\setlength\tabcolsep{0pt}
\begin{tabular*}{\textwidth}{@{\extracolsep{\fill}} l l r r r r }
\toprule
\multicolumn{1}{l}{Tasks} & \multicolumn{1}{l}{Datasets} & \multicolumn{1}{c}{Labels} & \multicolumn{1}{c}{Train} & \multicolumn{1}{c}{Valid} & \multicolumn{1}{r}{Test} \\
\midrule
Slot filling     & \text{ATIS \citep{hemphill-etal-1990-atis,dahl-etal-1994-expanding}} & 127 & 4,478 & 500 & 893\\
Slot filling    & \text{SNIPS \citep{coucke2018snips}} & 72 & 13,084 & 700 & 700\\
Chunking & \text{CoNLL 2000 \citep{tjong-kim-sang-buchholz-2000-introduction}} & 23 & 7,922 & 1,014 & 2,012\\
Named entity recognition & \text{OntoNotes 5.0, WSJ section \citep{ontonotes}} & 37 & 30,060 & 5,315 & 1,640\\
Parts-of-Speech tagging & \text{OntoNotes 5.0, WSJ section \citep{ontonotes}} & 48 & 30,060 & 5,315 & 1,640\\
Intent detection   & \text{ATIS \citep{hemphill-etal-1990-atis,dahl-etal-1994-expanding}} & 26 & 4,478 & 500 & 893\\
Intent detection  & \text{SNIPS \citep{coucke2018snips}} & 7 & 13,084 & 700 & 700\\
Sentiment classification  & \text{Positive/Negative \citep{10.1145/2783258.2783380}} & 2 & 2,100 & 300 & 600\\
Sentiment classification & \text{Pros/Cons \citep{ganapathibhotla-liu-2008-mining}} & 2 & 32,088 & 4,602 & 9,175\\
\bottomrule
\end{tabular*}
\caption{Tasks, datasets and their size.}
\end{table*}

\makeatletter
\setlength{\@fptop}{0pt}
\makeatother

\begin{table*}[tp!] \footnotesize
\captionsetup{singlelinecheck = false, justification=justified}
\setlength\tabcolsep{10pt}
\begin{tabular*}{\textwidth}{ l l  @{\hskip 2cm} c c c c c c}
\toprule
& \multicolumn{1}{l}{Tasks} & \multicolumn{1}{c}{Baseline} & \multicolumn{1}{c}{LT}
& \multicolumn{1}{c}{LT+R} & \bigcell{l}{LT+R \\ +CM} & \bigcell{c}{LT+R+ \\ CM+D1} & \bigcell{c}{LT+R+ \\ CM+D2}  \\
\midrule
\multirow{4}{*}{\rotatebox[origin=c]{90}{\textbf{F1}}}
& ATIS-Slot     & 96.93 & 95.88 & 89.67 & 96.07 & 96.72 & 94.60 \\
& SNIPS-Slot    & 91.36 & 90.02 & 58.91 & 83.41 & 86.44 & 87.81   \\
& Chunk    & 92.14 & 89.66 & 69.49 & 87.58 & 90.46 & 90.26   \\
& NER    & 86.43 & 81.87 & 51.09 & 67.50 & 78.54 & 82.47 \\
\midrule
\multirow{4}{*}{\rotatebox[origin=c]{90}{\textbf{Accuracy}}}
& PoS Tagging  & 96.34 & 95.72 & 90.02 & 94.86 & 95.84 & 95.88    \\
& ATIS-Intent  & 98.40 & 98.40 & 93.20 & 98.60 & 98.00 & 98.00   \\
& SNIPS-Intent & 99.43 & 98.86 & 89.43 & 99.00 & 99.00 & 99.00   \\
& Pos/Neg & 87.67 & 84.67 & 68.67 & 83.67 & 83.33 & 82.67  \\
& Pros/Cons   & 95.13 & 94.98 & 90.59 & 94.85 & 95.11 & 94.98   \\

\bottomrule
\end{tabular*}
\caption{Non-incremental performance of our models on validation sets for reproducibility purpose.}
\end{table*}

\begin{table*}[ht!] \footnotesize
\captionsetup{singlelinecheck = false, justification=justified}
\setlength\tabcolsep{0pt}
\begin{tabular*}{\textwidth}{@{\extracolsep{\fill}}l c c c c c c c}
\toprule
 \multicolumn{1}{l}{Tasks/Models} & \multicolumn{1}{l}{EO} & \multicolumn{1}{l}{CT} & \multicolumn{1}{l}{RC} & \multicolumn{1}{l}{EO$\Delta 1$} & \multicolumn{1}{l}{EO$\Delta2$} & \multicolumn{1}{l}{RC$\Delta1$} & \multicolumn{1}{l}{RC$\Delta2$}\\
\midrule
ATIS-Slot \\
\cmidrule{1-1}
Baseline & 0.029 & 0.012 & 0.963 & 0.009 & 0.003 & 0.986 & 0.995 \\
LT & 0.038 & 0.017 & 0.947 & 0.019 & 0.010 & 0.972 & 0.985 \\
LT+R & 0.000 & 0.000 & 1.000 & - & - & - & - \\
LT+R+CM & 0.000 & 0.000 & 1.000 & - & - & - & - \\
LT+R+CM+D1 & - & 0.000 & - & 0.000 & - & 1.000 & - \\
LT+R+CM+D2 & - & 0.000 & - & - & 0.000 & - & 1.000 \\
\midrule
SNIPS-Slot \\
\cmidrule{1-1}
Baseline & 0.147 & 0.078 & 0.805 & 0.067 & 0.038 & 0.906 & 0.945 \\
LT & 0.189 & 0.103 & 0.738 & 0.095 & 0.050 & 0.868 & 0.929\\
LT+R & 0.000 & 0.000 & 1.000 & - & - & - & - \\
LT+R+CM & 0.000 & 0.000 & 1.000 & - & - & - & - \\
LT+R+CM+D1 & - & 0.000 & - & 0.000 & - & 1.000 & - \\
LT+R+CM+D2 & - & 0.000 & - & - & 0.000 & - & 1.000 \\
\midrule
Chunk \\
\cmidrule{1-1}
Baseline & 0.166 & 0.046 & 0.743 & 0.073 & 0.057 & 0.852 & 0.874 \\
LT & 0.184 & 0.056 & 0.680 & 0.120 & 0.098 & 0.757 & 0.786\\
LT+R & 0.000 & 0.000 & 1.000 & - & - & - & - \\
LT+R+CM & 0.000 & 0.000 & 1.000 & - & - & - & - \\
LT+R+CM+D1 & - & 0.000 & - & 0.000 & - & 1.000 & - \\
LT+R+CM+D2 & - & 0.000 & - & - & 0.000 & - & 1.000 \\
\midrule
NER \\
\cmidrule{1-1}
Baseline & 0.072 & 0.019 & 0.898 & 0.037 & 0.023 & 0.935 & 0.952\\
LT & 0.078 & 0.022 & 0.883 & 0.040 & 0.025 & 0.926 & 0.944 \\
LT+R & 0.000 & 0.000 & 1.000 & - & - & - & - \\
LT+R+CM & 0.000 & 0.000 & 1.000 & - & - & - & - \\
LT+R+CM+D1 & - & 0.000 & - & 0.000 & - & 1.000 & - \\
LT+R+CM+D2 & - & 0.000 & - & - & 0.000 & - & 1.000 \\
\midrule
PoS Tagging \\
\cmidrule{1-1}
Baseline & 0.114 & 0.032 & 0.812 & 0.051 & 0.039 & 0.886 & 0.904\\
LT & 0.125 & 0.036 & 0.777 & 0.061 & 0.044 & 0.854 & 0.879\\
LT+R & 0.000 & 0.000 & 1.000 & - & - & - & - \\
LT+R+CM & 0.000 & 0.000 & 1.000 & - & - & - & - \\
LT+R+CM+D1 & - & 0.000 & - & 0.000 & - & 1.000 & - \\
LT+R+CM+D2 & - & 0.000 & - & - & 0.000 & - & 1.000 \\
\bottomrule
\end{tabular*}
\caption{Mean value of Edit Overhead, Correction Time Score and Relative Correctness on test sets for sequence tagging. $\Delta t$ denotes delay for $t$ time steps.}
\label{table:seqtagging} 
\vspace*{+4cm}
\end{table*}

\vfill

\makeatletter
\setlength{\@fptop}{0pt}
\makeatother

\begin{table*}[tp!] \footnotesize
\captionsetup{singlelinecheck = false, justification=justified}
\setlength\tabcolsep{0pt}
\begin{tabular*}{\textwidth}{@{\extracolsep{\fill}}l c c c c c c c}
\toprule
 \multicolumn{1}{l}{Tasks/Models} & \multicolumn{1}{l}{EO} & \multicolumn{1}{l}{CT} & \multicolumn{1}{l}{RC} & \multicolumn{1}{l}{EO$\Delta 1$} & \multicolumn{1}{l}{EO$\Delta2$} & \multicolumn{1}{l}{RC$\Delta1$} & \multicolumn{1}{l}{RC$\Delta2$}\\
\midrule
ATIS-Intent \\
\cmidrule{1-1}
Baseline & 0.488 & 0.238 & 0.812 & 0.333 & 0.238 & 0.874 & 0.923 \\
LT & 0.320 & 0.152 & 0.885 & 0.178 & 0.121 & 0.934 & 0.959 \\
LT+R & 0.276 & 0.191 & 0.836 & - & - & - & - \\
LT+R+CM & 0.174 & 0.097 & 0.925 & - & - & - & - \\
LT+R+CM+D1 & - & 0.056 & - & 0.097 & - & 0.958 & - \\
LT+R+CM+D2 & - & 0.032 & - & - & 0.073 & - & 0.976 \\
\midrule
SNIPS-Intent \\
\cmidrule{1-1}
Baseline & 0.294 & 0.176 & 0.857 & 0.202 & 0.145 & 0.896 & 0.925 \\
LT & 0.241 & 0.172 & 0.867 & 0.189 & 0.153 & 0.895 & 0.919 \\
LT+R & 0.120 & 0.131 & 0.883 & - & - & - & - \\
LT+R+CM & 0.188 & 0.112 & 0.915 & - & - & - & - \\
LT+R+CM+D1 & - & 0.073 & - & 0.130 & - & 0.944 & - \\
LT+R+CM+D2 & - & 0.044 & - & - & 0.083 & - & 0.969 \\
\midrule
Pos/Neg \\
\cmidrule{1-1}
Baseline & 0.358 & 0.249 & 0.829 & 0.266 & 0.220 & 0.859 & 0.881 \\
LT & 0.363 & 0.272 & 0.821 & 0.292 & 0.243 & 0.843 & 0.867\\
LT+R & 0.103 & 0.081 & 0.931 & - & - & - & - \\
LT+R+CM & 0.317 & 0.205 & 0.852 & - & - & - & - \\
LT+R+CM+D1 & - & 0.162 & - & 0.250 & - & 0.876 & - \\
LT+R+CM+D2 & - & 0.117 & - & - & 0.167 & - & 0.908 \\
\midrule
Pros/Cons \\
\cmidrule{1-1}
Baseline & 0.150 & 0.112 & 0.927 & 0.095 & 0.063 & 0.951 & 0.965 \\
LT & 0.158 & 0.113 & 0.925 & 0.092 & 0.061 & 0.952 & 0.966\\
LT+R & 0.095 & 0.071 & 0.943 & - & - & - & - \\
LT+R+CM & 0.098 & 0.069 & 0.952 & - & - & - & - \\
LT+R+CM+D1 & - & 0.040 & - & 0.059 & - & 0.972 & - \\
LT+R+CM+D2 & - & 0.027 & - & - & 0.042 & - & 0.981 \\
\bottomrule
\end{tabular*}
\caption{Mean value of Edit Overhead, Correction Time Score and Relative Correctness on test sets for sequence classification. $\Delta t$ denotes delay for $t$ time steps.}
\label{table:seqclassification} 
\vspace*{+11cm}
\end{table*}

\vfill
\null

\makeatletter
\setlength{\@fptop}{0pt}
\makeatother

\begin{table}[tp!] \small
\captionsetup{singlelinecheck = false, justification=justified}
\setlength\tabcolsep{0pt}
\begin{tabular*}{\columnwidth}{@{\extracolsep{\fill}}l c c c r}
\toprule
 \multicolumn{1}{l}{Tasks/Models} & \multicolumn{1}{c}{EO} & \multicolumn{1}{c}{CT} & \multicolumn{1}{c}{RC}\\
\midrule
ATIS-Slot \\
\cmidrule{1-1}
Baseline &  \\
\quad -- GloVe & 0.030 & 0.016 & 0.955 \\
\quad -- Pos & 0.109 & 0.058 & 0.836 \\
\quad -- GloVe \& Pos & 0.108 & 0.059 & 0.834 \\ 
LT+R+CM &  \\
\quad -- GloVe & 0.000 & 0.000 & 1.000 \\
\quad -- Pos & 0.000 & 0.000 & 1.000 \\
\quad -- GloVe \& Pos & 0.000 & 0.000 & 1.000 \\ 
\midrule
SNIPS-Slot \\
\cmidrule{1-1}
Baseline &  \\
\quad -- GloVe & 0.154 & 0.086 & 0.778 \\
\quad -- Pos & 0.264 & 0.172 & 0.597 \\
\quad -- GloVe \& Pos & 0.236 & 0.149 & 0.645 \\ 
LT+R+CM &  \\
\quad -- GloVe & 0.000 & 0.000 & 1.000\\
\quad -- Pos & 0.000 & 0.000 & 1.000\\
\quad -- GloVe \& Pos & 0.000 & 0.000 & 1.000 \\ 
\midrule
Chunk \\
\cmidrule{1-1}
Baseline &  \\
\quad -- GloVe & 0.177 & 0.059 & 0.675 \\
\quad -- Pos & 0.325 & 0.190 & 0.296 \\
\quad -- GloVe \& Pos & 0.360 & 0.200 & 0.276 \\ 
LT+R+CM &  \\
\quad -- GloVe & 0.000 & 0.000 & 1.000 \\
\quad -- Pos & 0.000 & 0.000 & 1.000 \\
\quad -- GloVe \& Pos & 0.000 & 0.000 & 1.000 \\ 
\midrule
NER \\
\cmidrule{1-1}
Baseline &  \\
\quad -- GloVe & 0.079 & 0.021 & 0.875 \\
\quad -- Pos & 0.096 & 0.040 & 0.792 \\
\quad -- GloVe \& Pos & 0.114 & 0.046 & 0.759 \\ 
LT+R+CM &  \\
\quad -- GloVe & 0.000 & 0.000 & 1.000 \\
\quad -- Pos & 0.000 & 0.000 & 1.000 \\
\quad -- GloVe \& Pos & 0.000 & 0.000 & 1.000 \\ 
\midrule
PoS Tagging \\
\cmidrule{1-1}
Baseline &  \\
\quad -- GloVe & 0.124 & 0.035 & 0.790 \\
\quad -- Pos & 0.184 & 0.079 & 0.567 \\
\quad -- GloVe \& Pos & 0.195 & 0.087 & 0.545 \\ 
LT+R+CM &  \\
\quad -- GloVe & 0.000 & 0.000 & 1.000 \\
\quad -- Pos & 0.000 & 0.000 & 1.000\\
\quad -- GloVe \& Pos & 0.000 & 0.000 & 1.000 \\ 
\bottomrule
\end{tabular*}
\caption{Ablation of GloVe and positional embeddings on the baseline and LT+R+CM for incremental metrics in sequence tagging.}
\end{table}

\vfill
\null

\begin{table}[t] \small
\captionsetup{singlelinecheck = false, justification=justified}
\setlength\tabcolsep{0pt}
\begin{tabular*}{\columnwidth}{@{\extracolsep{\fill}}l c c c r}
\toprule
 \multicolumn{1}{l}{Tasks/Models} & \multicolumn{1}{c}{EO} & \multicolumn{1}{c}{CT} & \multicolumn{1}{c}{RC}\\
\midrule
ATIS-Intent \\
\cmidrule{1-1}
Baseline &  \\
\quad -- GloVe & 0.505 & 0.248 & 0.798 \\
\quad -- Pos & 0.472 & 0.235 & 0.807 \\
\quad -- GloVe \& Pos & 0.497 & 0.239 & 0.802 \\ 
LT+R+CM &  \\
\quad -- GloVe & 0.179 & 0.101 & 0.922 \\
\quad -- Pos & 0.166 & 0.091 & 0.927 \\
\quad -- GloVe \& Pos & 0.190 & 0.115 & 0.916 \\ 
\midrule
SNIPS-Intent \\
\cmidrule{1-1}
Baseline &  \\
\quad -- GloVe & 0.247 & 0.172 & 0.869 \\
\quad -- Pos & 0.263 & 0.165 & 0.872 \\
\quad -- GloVe \& Pos & 0.195 & 0.139 & 0.892 \\ 
LT+R+CM &  \\
\quad -- GloVe & 0.162 & 0.093 & 0.925 \\
\quad -- Pos & 0.179 & 0.105 & 0.921 \\
\quad -- GloVe \& Pos & 0.176 & 0.102 & 0.919 \\ 
\midrule
Pos/Neg \\
\cmidrule{1-1}
Baseline &  \\
\quad -- GloVe & 0.452 & 0.368 & 0.769 \\
\quad -- Pos & 0.356 & 0.256 & 0.826 \\
\quad -- GloVe \& Pos & 0.477 & 0.375 & 0.774 \\ 
LT+R+CM &  \\
\quad -- GloVe & 0.324 & 0.197 & 0.859 \\
\quad -- Pos & 0.284 & 0.192 & 0.856 \\
\quad -- GloVe \& Pos & 0.306 & 0.180 & 0.865 \\ 
\midrule
Pros/Cons \\
\cmidrule{1-1}
Baseline &  \\
\quad -- GloVe & 0.162 & 0.116 & 0.924 \\
\quad -- Pos & 0.167 & 0.120 & 0.924 \\
\quad -- GloVe \& Pos & 0.188 & 0.142 & 0.910 \\ 
LT+R+CM &  \\
\quad -- GloVe & 0.096 & 0.065 & 0.955 \\
\quad -- Pos & 0.095 & 0.064 & 0.955 \\
\quad -- GloVe \& Pos & 0.094 & 0.066 & 0.954 \\ 
\bottomrule
\end{tabular*}
\caption{Ablation of GloVe and positional embeddings on the baseline and LT+R+CM for incremental metrics in sequence classification.}
\end{table}

\end{document}